# Multiple Kernel Learning from Noisy Labels by Stochastic Programming

**Tianbao Yang**[†]      YANGTIA1@MSU.EDU
**Mehrdad Mahdavi**[†]      MAHDAVIM@MSU.EDU
**Rong Jin**[†]      RONGJIN@MSU.EDU
**Lijun Zhang**[‡]      ZLJZJU@ZJU.EDU.CN
**Yang Zhou**[∗]      YOUNG.ZHOU@GMAIL.COM

[†]Department of Computer Science and Engineering, Michigan State University, East Lansing, MI 48824, USA
[‡]Zhejiang Key Laboratory of Service Robot, College of Computer Science, Zhejiang University, Hangzhou, China
[∗]Yahoo Labs!, Santa Clara, CA 95054, USA

## Abstract

We study the problem of multiple kernel learning from noisy labels. This is in contrast to most of the previous studies on multiple kernel learning that mainly focus on developing efficient algorithms and assume perfectly labeled training examples. Directly applying the existing multiple kernel learning algorithms to noisily labeled examples often leads to suboptimal performance due to the incorrect class assignments. We address this challenge by casting multiple kernel learning from noisy labels into a stochastic programming problem, and presenting a minimax formulation. We develop an efficient algorithm for solving the related convex-concave optimization problem with a fast convergence rate of $O(1/T)$ where $T$ is the number of iterations. Empirical studies on UCI data sets verify both the effectiveness and the efficiency of the proposed algorithm.

## 1. Introduction

Multiple Kernel Learning (MKL) (Lanckriet et al., 2004) has attracted a significant amount of interest in both machine learning and data mining communities due to the success of kernel methods (Schölkopf & Smola, 2001). MKL aims to learn an optimal combination of multiple kernels and a nonlinear classifier from the Reproducing Kernel Hilbert Space (RKHS) endowed with the combined kernel.

Most of the previous studies on MKL has focused on designing efficient algorithms for solving the related optimization problem. Research has also been done to study the effect of different regularizers on the combination of multiple kernels, including sparse regularizer $\ell_1$ norm, non-sparse regularizer $\ell_2$ norm, and in general $\ell_p$ norm. One limitation of these studies is that they all assume perfectly labeled training examples, which significantly limits their application to problems where class assignments are often noisy and inaccurate. Noisy class assignments could arise either from the biases of human subjects or because the class labels are automatically derived from side information (e.g., hyperlink information (Yang et al., 2010)).

In this work, we address this limitation by casting MKL from noisy labels into a stochastic programming problem (Kall & Wallace, 1994). The key idea is to introduce a binary random variable for each training example to indicate if the class assignment of the example is correct. Using introduced binary random variables, we turn the deterministic constraint in MKL into a chance constraint (Ben-Tal et al., 2009), leading to a stochastic programming formulation[1]. By assuming that the percentage of incorrectly labeled training examples is given, we approximate the stochastic programming problem into a convex-concave optimization problem. Unlike many previous studies (Lawrence & Schölkopf, 2001; Pal et al., 2007; Yang et al., 2010) on learning with noisy labeled data that make strong assumptions about the noise model (e.g. conditional independence assumption between noisy label and the data given the true label), our framework depends only on a mild assumption about the noise (see section 3.2), making it practically more useful. Notably, we do NOT assume the label noise of different examples are independent, a common assumption shared by most existing studies on learning from noisy labels (Biggio et al., 2011; Yang et al., 2010). Based on the mirror prox method (Nemirovski, 2005), we develop a first order method for solving the related convex-concave optimization problem. Our analysis shows the convergence rate of $O(1/T)$ for the proposed algorithm, significantly faster than the classical $O(1/\sqrt{T})$ rate. Empirical studies on five UCI data sets confirm the effectiveness and the efficiency of the proposed framework and the optimization algorithm.

---

[1]It is important to distinguish stochastic programming (Kall & Wallace, 1994) from stochastic optimization (Robbins & Monro, 1951). The former refers to the set of problems where the solution is affected by the uncertainty of the system to be optimized, while the later refers to the optimization algorithm (usually iterative) that depends on random variables. In particular, stochastic optimization can be applied to solving a deterministic optimization problem.

## 2. Related Work

Our work is closely related to MKL. Various criteria have been developed to find the optimal kernel combination, such as maximum classification margin (Lanckriet et al., 2004) and kernel alignment (Cristianini et al., 2002; Cortes et al., 2010a). Among them, maximum margin MKL receives most attention due to its empirical success and its close relationship with Support Vector Machines (SVMs). Many algorithms have been developed for max-margin MKL by formulating the problem into Semi-Definite Programming (Lanckriet et al., 2004), Second Order Cone Programming (Bach et al., 2004), and Semi-Infinite Linear Programming (Sonnenburg et al., 2006). Due to their high computational cost, these approaches are unable to handle large data sets and a large number of kernels. A number of efficient algorithms, based on alternating optimization, have been proposed for MKL (Rakotomamonjy et al., 2008; Xu et al., 2010). Besides efficient learning algorithms, various regularizers have been studied for MKL, including $\ell_1$ norm (Rakotomamonjy et al., 2008; Xu et al., 2010), $\ell_2$ norm (Cortes et al., 2009), and $\ell_p$ norm (Kloft et al., 2009).

Our work is also related to learning with noisy labels. Lawrence & Schölkopf (2001) propose a probabilistic model for learning a kernel Fisher Discriminant from noisy labels. Pal et al. (2007) present a probabilistic model for extracting location information for events with noisy training labels. Ramakrishnan et al. (2005) propose a Bayeisan model for learning with approximate, noisy or incomplete labels. Yang et al. (2010) propose a generalized maximum entropy model for learning from noisy side information. These probabilistic approaches have to make strong assumptions about label noise, which significantly limit their applications to real-world problems. In addition, it is difficult to adapt them to MKL. Several recent studies (Huang et al., 2010) address the limitation of probabilistic approaches by exploring the robust optimization (Ben-Tal et al., 2009). Our study is particularly related to the recent work on robust SVM (Xu et al., 2006; Yu et al., 2011) in which a SVM classifier is learned in the presence of outliers. Our work differs from these studies in two aspects. First, we address a different learning problem (i.e., MKL from noisy labels). Second, our work is based on stochastic programming that makes least possible assumption about the noise model compared to the other approaches.

## 3. Multiple Kernel Learning From Noisy Labels

We first review a formulation of MKL based on the equivalence between MKL and group Lasso (Xu et al., 2010; Bach, 2008). We then describe the problem of MKL from noisy labels and present a stochastic programming framework to formulate it. Finally, we present an efficient algorithm for solving the related convex-concave optimization problem.

### 3.1. Multiple Kernel Learning (MKL)

Let $\mathcal{D} = \{(\mathbf{x}_i, y_i), i = 1, \cdots, n\}$ be the training data, where $\mathbf{x}_i \in \mathbb{R}^d$ denotes the $i$th instance and $y_i \in \{-1, 1\}$ denotes its binary label. We use $\mathbf{y} = (y_1, \cdots, y_n)^\top$ to represent the class assignment of all training examples. We denote by $\{\kappa_j(\cdot, \cdot) : \mathbb{R}^d \times \mathbb{R}^d \mapsto \mathbb{R}, j \in [m]\}$ the set of $m$ kernels to be combined, and by $\mathcal{H}_{\kappa_j}$ the corresponding Reproducing Kernel Hilbert Space (RKHS) endowed by $\kappa_j$. We use $\mathbf{u} = (u_1, \cdots, u_m)^\top$ for the combination weights of multiple kernels, $\kappa_{\mathbf{u}} = \sum_j u_j \kappa_j$ for the combined kernel, and $\mathcal{H}_{\kappa_{\mathbf{u}}}$ for the RKHS endowed by $\kappa_{\mathbf{u}}$. In this work, we consider $\mathbf{u} \in \Delta$, where $\Delta = \{\mathbf{u} \in \mathbb{R}_+^m : \sum_j u_j = 1\}$ is a simplex. MKL can be cast as the following problem:

$$\min_{f \in \mathcal{H}_{\kappa_{\mathbf{u}}}, \mathbf{u} \in \Delta} \frac{\lambda}{2} \|f\|^2_{\mathcal{H}_{\kappa_{\mathbf{u}}}} + \frac{1}{n} \sum_{i=1}^n \ell(y_i f(\mathbf{x}_i)), \quad (1)$$

where $\ell(z) = \max(0, 1-z)$ is the hinge loss function. It has been shown that the problem in (1) is equivalent to the following problem (Micchelli & Pontil, 2005),

$$\min_{\{f_j\}_{j=1}^m} \frac{\lambda}{2} \left(\sum_{j=1}^m \|f_j\|_{\mathcal{H}_{\kappa_j}}\right)^2 + \frac{1}{n}\sum_{i=1}^n \ell\left(y_i \sum_{j=1}^m f_j(\mathbf{x}_i)\right), \tag{2}$$

or equivalently

$$\min_{t,\{f_j\}_{j=1}^m} \frac{\lambda}{2}\left(\sum_{j=1}^m \|f_j\|_{\mathcal{H}_{\kappa_j}}\right)^2 + t \tag{3}$$

$$\text{s.t.} \quad \frac{1}{n}\sum_{i=1}^n \ell\left(y_i \sum_{j=1}^m f_j(\mathbf{x}_i)\right) \leq t, \tag{4}$$

where $f_j$ belongs to $\mathcal{H}_{\kappa_j}$ and $t$ is a slack variable to be optimized. Given the solutions $f_j, j \in [m]$ to (3), the final classifier is defined as $f(\mathbf{x}) = \sum_{j=1}^m f_j(\mathbf{x})$. In the sequel, we use the notation $f(\mathbf{x}) = \sum_{j=1}^m f_j(\mathbf{x})$ to simplify our presentation. Our formulation for MKL from noisy labels is based on the formulation in (3). In the next subsection, we present a stochastic programming framework for MKL from noisy labels. Due to the limit of space, we put the proofs of most analysis in the supplementary material.

### 3.2. A Stochastic Programming Framework for MKL from Noisy Labels

In the case of noisy labels, we have some incorrect class assignments for the training examples in $\mathcal{D}$. The key challenge is that we do not know which examples are incorrectly labeled. To facilitate learning from noisy labels, we assume the noise level of class assignments $q \in [0, 1/2)$, i.e., the expected probability for any randomly chosen example to be incorrectly labeled, is given.

For a given pair $(\mathbf{x}, y)$, let $\xi(\mathbf{x}, y) \in \{0, 1\}$ be a binary random variable indicating if $y$ is a correct label of $\mathbf{x}$ (1) or not (0), and $p(\mathbf{x}, y) = \mathrm{E}_{\xi|(\mathbf{x},y)}[\xi(\mathbf{x}, y)]$ be the probability for $y$ to be a correct label of $\mathbf{x}$.

**Assumption 1.** *The noise level $q$ is given, i.e.,*

$$q = 1 - \mathrm{E}_{(\mathbf{x},y)}[p(\mathbf{x}, y)].$$

With Assumption 1, we present the following proposition to bound the empirical mean of $p(\mathbf{x}, y)$ on the training examples.

**Proposition 1.** *Let $\xi_i, i \in [n]$ denote the binary indicator variable of noise on the training examples, and $p_i = \mathrm{E}[\xi_i]$. Given the noise level $q$, with probability at least $1 - \epsilon$, we have*

$$\frac{1}{n}\sum_{i=1}^n p_i \leq 1 - q + \frac{\tau}{\sqrt{n}},$$

where $\tau = \sqrt{(1/2)\ln(1/\epsilon)}$.

The proposition follows directly from the Hoeffding's inequality (Boucheron et al., 2004).

To handle the noisy labels, we consider a stochastic programming (Kall & Wallace, 1994) framework. More specifically, given the unknown random variables $\xi = (\xi_1, \ldots, \xi_n)$, where $\xi_i = \xi(\mathbf{x}_i, y_i)$, we relax the deterministic constraint in (4) into a chance constraint (Ben-Tal et al., 2009)

$$\Pr\left(\frac{1}{n}\sum_{i=1}^n \xi_i \ell\left(y_i f(\mathbf{x}_i)\right) > t\right) \leq \epsilon, \tag{5}$$

where $\Pr(\cdot)$ takes over the unknown joint distribution of binary random variables $\xi$, and $\epsilon \in (0, 1)$ bounds the probability for the constraint in (3) to be violated. The chance constraint in (5) requires that there is only a small chance for the constraint to be violated by the unknown correctly labeled examples. It has also been used for handling the uncertainty before. In (Bhadra et al., 2010), the authors introduce the chance constraints to handle the uncertainty in the elements of a kernel matrix, while the chance constraint in (5) is introduced to handle the uncertainty in the class labels.

Using the chance constraint in (5), we have the following stochastic programming problem for MKL from noisy labels:

$$\min_{t,\{f_j\}_{j=1}^m} \frac{\lambda}{2}\left(\sum_{j=1}^m \|f_j\|_{\mathcal{H}_{\kappa_j}}\right)^2 + t \tag{6}$$

$$\text{s.t.} \quad \Pr\left(\frac{1}{n}\sum_{i=1}^n \xi_i \ell\left(y_i f(\mathbf{x}_i)\right) > t\right) \leq \epsilon.$$

A major challenge of solving the stochastic programming problem in (6) is that the joint distribution for $\xi$ is unknown. The following lemma allows us to alleviate this challenge.

**Lemma 1.** *If the following inequality holds,*

$$\frac{1}{n}\sum_{i=1}^n \mathrm{E}[\xi_i]\ell\left(y_i f(\mathbf{x}_i)\right) \leq t - \tau\sqrt{\frac{1}{n^2}\sum_{i=1}^n \ell^2\left(y_i f(\mathbf{x}_i)\right)},$$

*where $\tau = \sqrt{(1/2)\ln(1/\epsilon)}$, then the chance constraint in (5) is satisfied.*

The proof follows immediately from the McDiarmid inequality (Boucheron et al., 2004). It is important to note that Lemma 1 holds WITHOUT having to assume that the binary random variables $\{\xi_i\}_{i=1}^m$ are independent, a common assumption that appears in almost all the studies on learning from noisy labels.

Using Lemma 1, we relax the problem in (6) into the following optimization problem

$$\min_{t,\{f_j\}_{j=1}^m} \quad \frac{\lambda}{2}\left(\sum_{j=1}^m \|f_j\|_{\mathcal{H}_{\kappa_j}}\right)^2 + t \quad (7)$$

$$\text{s.t.} \quad \frac{1}{n}\sum_{i=1}^n p_i \ell\left(y_i f(\mathbf{x}_i)\right) + \frac{\tau}{n}\sqrt{\sum_{i=1}^n \ell^2\left(y_i f(\mathbf{x}_i)\right)} \le t, \quad (8)$$

where $p_i$ denotes $\mathrm{E}[\xi_i]$. We can turn the constrained problem into non-constrained problem by replacing $t$ in (7) with the lower bound in (8). There are two problems with directly optimizing (7). First, the second term in the lower bound of $t$ in (8) is square-root of a quadratic form on the training loss, making the optimization problem difficult to solve. Second, the variables $\{p_i\}_{i=1}^n$ are unknown. Without knowing the value of $\{p_i\}_{i=1}^n$, it is impossible to solve the optimization problem in (7).

To address the first problem, we use the inequality $\sqrt{\sum_{i=1}^n \ell_i^2} \le \sum_{i=1}^n |\ell_i|$ to relax the constraint in (8) into

$$\frac{1}{n}\sum_{i=1}^n p_i \ell\left(y_i f(\mathbf{x}_i)\right) + \frac{\tau}{n}\sum_{i=1}^n \ell\left(y_i f(\mathbf{x}_i)\right) \le t. \quad (9)$$

Note that inequality (9) indicates inequality (8), and therefore guarantees that the chance constraint in (5) holds. Then we turn problem (7) into

$$\min_{t,\{f_j\}_{j=1}^m} \quad \frac{\lambda}{2}\left(\sum_{j=1}^m \|f_j\|_{\mathcal{H}_{\kappa_j}}\right)^2 + t$$

$$\text{s.t.} \quad \frac{1}{n}\sum_{i=1}^n (p_i + \tau)\ell\left(y_i f(\mathbf{x}_i)\right) \le t,$$

or equivalently

$$\min_{\{f_j\}_{j=1}^m} \quad \frac{\lambda}{2}\left[\sum_{j=1}^m \|f_j\|_{\mathcal{H}_{\kappa_j}}\right]^2 + \frac{1}{n}\sum_{i=1}^n (p_i + \tau)\ell\left(y_i f(\mathbf{x}_i)\right).$$

To address the second problem, we propose the following minimax formulation

$$\min_{\{f_j\}_{j=1}^m} \max_{\mathbf{p} \in \mathcal{Q}} \frac{\lambda}{2}\left[\sum_{j=1}^m \|f_j\|_{\mathcal{H}_{\kappa_j}}\right]^2 + \frac{1}{n}\sum_{i=1}^n (p_i + \tau)\ell\left(y_i f(\mathbf{x}_i)\right), \quad (10)$$

where $\mathcal{Q} = \{\mathbf{p} \in [0,1]^n, \sum_i p_i \le (1-q)n + \tau\sqrt{n}\}$ is a domain for $\mathbf{p}$ justified by Proposition 1.

**Remark:** We choose to maximize over $\mathbf{p} \in \mathcal{Q}$ because it guarantees that the loss of any choice of correctly labeled examples is minimized. The idea of using the worst case analysis is closely related to robust optimization (Ben-Tal et al., 2009), which has been adopted by several recent studies, including budget SVM (Dekel & Singer, 2006), and robust metric learning (Huang et al., 2010). Note that an alternative approach is to consider the best case analysis (a strategy taken in robust SVM (Xu et al., 2006; Yu et al., 2011)) by minimizing the robust hinge loss, which can be defined as $\min_{\mathbf{p}\in\mathcal{Q}} \sum_i (p_i \ell_i + 1 - p_i)$, where $\ell_i$ denotes the loss on $i$th example, to address the uncertainty arising from noisy labels.

There are several problems with the alternative approach. First, minimization over $\mathbf{p}$ will lead to a non-convex optimization problem, as shown in (Xu et al., 2006), making it difficult, if not impossible, to develop an efficient learning algorithm to find the global optimum. Second, unless a strong assumption is made about the examples with incorrect labels, minimization over $\mathbf{p}$ will not provide any guarantee about the generalized performance of the resulting classifier. Third, the formulation with minimization over $\mathbf{p}$ does not reduce to the normal case in (2) when there is no noise.

In contrast, our problem is a convex-concave problem, which allows us to derive an efficient optimization algorithm to solve it. Also, we do NOT make *any* assumption on the noisy labels except assuming that the noise level is given. More ever, the generalization error of the kernel classifier learned from (10) is given in Theorem 1, which also justifies the maximization over $\mathbf{p}$. Finally, it is straightforward to show that our formulation in (10) reduces to (2) when there is no noise. This point is also demonstrated by our empirical studies.

**Theorem 1.** *Assume that the number of incorrectly labeled instances in $\mathcal{D}$ is no more than $nq$. Let $f_j, j \in [m]$ be the final solutions to (10) and set $f(\mathbf{x}) = \sum_{j=1}^m f_j(\mathbf{x})$. With probability at least $1 - \delta$, we have*

$$\mathrm{E}_{(\mathbf{x},y)}\left[\ell\left(yf(\mathbf{x})\right)\right] \le \max_{\mathbf{p}\in\mathcal{Q}} \frac{1}{n(1-q)}\sum_{i=1}^n p_i \ell\left(y_i f(\mathbf{x}_i)\right)$$

$$+ \sqrt{\frac{2(1+\tau)}{n(1-q)\lambda}}\left[4 + \sqrt{\left(1 + \frac{\lambda}{2(1+\tau)}\right)\ln\frac{1}{\delta}}\right].$$

**Remark:** The bound scales with $1/(1-q)$, so when the noise level $q$ is large, the generalization bound is also large. Additionally, we can see that with a probability $(1 - m^{-k})$, where $k$ is an integer, the generaliza-

Multiple Kernel Learning From Noisy Labels

tion error bound has an additional term of $\sqrt{\ln m/n}$, which is a tight bound for $\ell_1$-regularized MKL in terms of the number of kernels (Ying & Campbell, 2009; Cortes et al., 2010b).

### 3.3. An Efficient Optimization Algorithm

In this section, we present an efficient algorithm for solving the minimax problem in (10). We first present an alternative formulation for (10), i.e.,

$$\min_{\{f_j\}_{j=1}^m} \max_{\alpha \in \mathcal{Q}_\alpha} \frac{\lambda}{2} \left( \sum_{j=1}^m \|f_j\|_{\mathcal{H}_{\kappa_j}} \right)^2 + \frac{1}{n} \sum_{i=1}^n \alpha_i \left(1 - y_i f(\mathbf{x}_i)\right), \quad (11)$$

where $\mathcal{Q}_\alpha = \{\alpha : \alpha \in [0, 1+\tau]^n, \|\alpha\|_1 \leq \rho\}$, and $\rho = (1 - q + \tau + \tau/\sqrt{n})n$. This is obtained by writing $\ell(y_i f(\mathbf{x}_i)) = \max(0, 1 - y_i f(\mathbf{x}_i)) = \max_{\beta_i \in [0,1]} \beta_i (1 - y_i f(\mathbf{x}_i))$, and introducing $\{\alpha_i = \beta_i(p_i + \tau)\}_{i=1}^n$.

Before presenting the optimization algorithm, we introduce a few notations that will be used throughout this section. We denote by $f = (f_1, \cdots, f_m)^\top \in \mathcal{H}$ where $\mathcal{H} = (\mathcal{H}_{\kappa_1}, \cdots, \mathcal{H}_{\kappa_m})$, and by $R(f)$ and $L(f, \alpha)$ the first term and the second term in the objective function in (11), respectively. We write the problem in (11) as

$$\min_{f \in \mathcal{H}} \max_{\alpha \in \mathcal{Q}_\alpha} F(f, \alpha) = R(f) + L(f, \alpha).$$

In the following, we refer to $f$ as the primal variable and $\alpha$ as the dual variable. We denote by $\nabla_f L(f, \alpha) = (\nabla_{f_1} L(f, \alpha), \cdots, \nabla_{f_m} L(f, \alpha))^\top$ and $\nabla_\alpha L(f, \alpha)$ the partial gradients of $L(f, \alpha)$ in terms of $f$ and $\alpha$, respectively. We use the notations $\|f_j\| = \|f_j\|_{\mathcal{H}_{\kappa_j}}$, and $\|f\|^2 = \sum_j \|f_j\|_{\mathcal{H}_{\kappa_j}}^2$ for short. We denote by $\prod_{\mathcal{Q}}[\hat{\mathbf{v}}]$ the projection $\hat{\mathbf{v}}$ into domain $\mathcal{Q}$, i.e. $\prod_{\mathcal{Q}}[\hat{\mathbf{v}}] = \arg\min_{\mathbf{v} \in \mathcal{Q}} \frac{1}{2}\|\mathbf{v} - \hat{\mathbf{v}}\|^2$, where $\|\cdot\|$ is $\ell_2$ norm when it is applied to a vector, and a RKHS norm when applied to a function.

Next, we present an accelerated mirror prox method that extends the mirror prox method (Nemirovski, 2005) to efficiently solve the convex-concave optimization problem in (11). The main limitation of the original mirror prox method is that it is only applicable to smooth objective functions whose gradients are Lipschitz continuous, which unfortunately is not the case for the problem in (11) because $R(f)$ is not a smooth function in $f$. Algorithm 1 outlines the key steps of this method. In Algorithm 1, we maintain two copies for the dual variables (i.e., $\alpha$ and $\beta$), but only one copy for the primal variable $f$. This is in contrast to the mirror prox method that introduces two copies for both primal and dual variables. Another key difference

---

**Algorithm 1** An Accelerated Mirror Prox Method

1: **Input**: step size $\gamma = \sqrt{n/(2m)}$
2: **Initialization**: $\beta_0 = 0, f^0 = 0$
3: **for** $t = 1, 2, \ldots, T$ **do**
4: $\quad \alpha_t = \prod_{\mathcal{Q}_\alpha}[\beta_{t-1} + \gamma \nabla_\alpha L(f^{t-1}, \beta_{t-1})]$
5: $\quad f^t = \arg\min_{f \in \mathcal{H}} \frac{1}{2}\left\|f - \widehat{f}^{t-1}\right\|^2 + \gamma R(f)$
$\quad\quad$ where $\widehat{f}^{t-1} = f^{t-1} - \gamma \nabla_f L(f^{t-1}, \alpha_t)$
6: $\quad \beta_t = \prod_{\mathcal{Q}_\alpha}[\beta_{t-1} + \gamma \nabla_\alpha L(f^t, \alpha_t)]$
7: **end for**
8: **Output**: $\widehat{f}_T = \sum_t f^t/T, \widehat{\alpha}_T = \sum_t \alpha_t/T$

---

between Algorithm 1 and the mirror prox method is that in step 5, we update the primal variable $f$ by a composite gradient mapping (Nesterov, 2007), instead of a gradient mapping. It is this step that allows us to solve the convex-concave optimization problems efficiently with a convergence rate of $O(1/T)$, without having to assume that the gradients of the objective function are Lipschitz continuous. It is important to point out that accelerated proximal gradient method by Tseng (Tseng, 2008) is not applicable to (11), since it requires the Lipschitz continuous gradients.

In order to efficiently implement Algorithm 1, we need to efficiently solve the optimization problems in steps 4, 5, and 6. The gradient mapping problems in steps 4 and 6 can be solved by utilizing the following lemma.

**Lemma 2.** *The optimal solution $\alpha^*$ to $\prod_{\mathcal{Q}_\alpha}[\widehat{\alpha}]$ is given by $\alpha_i^* = [\widehat{\alpha}_i - \eta]_{[0, 1+\tau]}, i = 1, \cdots, n$, where $[s]_{[a,b]}$ is projection of the number $s$ into the range $[a, b]$, and $\eta = 0$ if $\sum_i [\widehat{\alpha}_i]_{[0, 1+\tau]} < \rho$, otherwise $\eta$ is the solution to the following equation*

$$\sum_i [\widehat{\alpha}_i - \eta]_{[0, 1+\tau]} - \rho = 0. \quad (12)$$

Since $\sum_i [\widehat{\alpha}_i - \eta]_{[0, 1+\tau]} - \rho$ is monotonically decreasing function in $\eta$, we can efficiently compute $\eta$ in (12) by bisection search.

The composite gradient mapping in step 5 is

$$\min_{\{f_j\}_{j=1}^m} \frac{1}{2} \sum_j \|f_j - \widehat{f}_j^{t-1}\|^2 + \frac{\gamma\lambda}{2} \left( \sum_{j=1}^m \|f_j\| \right)^2, \quad (13)$$

where $\widehat{f}_j^{t-1} = f_j^{t-1} - \gamma \nabla_{f_j} L(f^{t-1}, \alpha_t)$, and it can be solved by the following lemma.

**Lemma 3.** *The optimal solution to (13) denoted by*

$f_j^*, j \in [m]$ is given by

$$f_j^* = \left[1 - \frac{\gamma\lambda\mu_t}{2\|\widehat{f}_j^{t-1}\|}\right]_+ \widehat{f}_j^{t-1},$$

where $[z]_+ = z$ if $z > 0$ and $0$ otherwise, and $\mu_t$ is the solution to the following equation,

$$\sum_j \left[1 - \frac{\gamma\lambda\mu_t}{2\|\widehat{f}_j^{t-1}\|}\right]_+ \|\widehat{f}_j^{t-1}\| - \mu_t = 0,$$

where $\mu_t$ can be computed by efficient bisection search.

**Remark:** Note that since the partial gradient $\nabla_{f_j} L(f, \alpha) = -(1/n)\sum_{i=1}^n \alpha_i y_i \kappa_j(\mathbf{x}_i, \cdot)$, hence, when updating the kernel classifier $f_j$, we can write it in a parameterized form $f_j = \sum_{i=1}^n z_i y_i \kappa_j(\mathbf{x}_i, \cdot)$ and update the coefficients $\mathbf{z} = (z_1, \cdots, z_n)$ at each iteration.

We conclude this subsection by presenting the convergence rate for Algorithm 1.

**Theorem 2.** *By running Algorithm 1 with $T$ iterations, we have*

$$\max_{\alpha \in \mathcal{Q}_\alpha} F(\widehat{f}_T, \alpha) - \min_f F(f, \widehat{\alpha}_T) \leq \frac{(\|f^*\|^2 + \|\alpha^*\|_2^2)\sqrt{m}}{\sqrt{2n}T},$$

*where $f^* = (f_1^*, \cdots, f_m^*)^\top = \arg\min_f F(f, \widehat{\alpha}_T)$ and $\alpha^* = \arg\max_{\alpha \in \mathcal{Q}_\alpha} F(\widehat{f}_T, \alpha)$.*

Theorem 2 indicates a $O(1/T)$ convergence rate for the accelerated mirror prox method presented in Algorithm 1 that is significantly faster than traditional $O(1/\sqrt{T})$ convergence rate for non-smooth optimization problems. This is also demonstrated by our empirical studies.

## 4. Experiments

In this section, we simulate our experiments on UCI data sets to verify the effectiveness and efficiency of the proposed algorithm for MKL from noisy labels. In a simulated environment, we can control the noise level to better understand the behavior of the proposed algorithm under different noise levels compared to baseline algorithms. We choose five data sets from UCI repository that have been used in MKL studies (Rakotomamonjy et al., 2008; Xu et al., 2010). The statistics of the data sets are summarized in Table 1. We normalize the data by scaling each attribute to $[0, 1]$. This is done by first subtracting each attribute from its minimal value and then dividing it by the difference between the maximal and the minimal value of the attribute. To generate label noise, we randomly flip the class label of each example with a probability of $q$. To create multiple kernels, we follow the setup in (Xu et al., 2010) to generate Gaussian kernels with 10 different width $\{2^{-3}, 2^{-2}, \cdots, 2^6\}$ for all features as well as for individual features, leading to a total of $m = 10(d+1)$ kernels for each data set, where $d$ is the number of features. We split the data into 80% for training and 20% for testing.

In the experiments, we focus on verifying the proposed stochastic programming framework for handling the noise in labels. We choose two baselines for comparison where one directly optimizes the objective in (2) assuming the labels are all correct, and the other one adopts a different strategy (i.e. minimization instead of maximization over $\mathbf{p}$) to handle the noise. By comparing with the first baseline, we are able to verify that existence of noise in labels significantly deteriorates the performance and therefore handling the noise is important. By comparing with the second baseline, we are able to verify the proposed stochastic programming framework with maximization over $\mathbf{p}$ is a good choice for handling the noise. For the first baseline, we choose Simple MKL (**SiPMKL**) algorithm, a state-of-the-art algorithm for $\ell_1$ regularized MKL[2]. For the second baseline, we extend the idea of robust SVM (Xu et al., 2006) to MKL from noisy labels by using the robust hinge loss and minimizing over $\mathbf{p} \in \mathcal{Q}$. We refer to this baseline as **MiPMKL**. Finally, we refer to proposed algorithm as **StPMKL**.

In implementing the proposed algorithm, we project $\alpha_i$ into $[0, 1]$ by absorbing the upper bound $1 + \tau$ into the regularization parameter $\lambda$ and the bound parameter $\rho(\tau)$. The regularization parameter $\lambda$ in the proposed algorithm (and baselines as well) is chosen by cross validation on a validation data of 10% examples randomly selected from the training data. The parameter $\rho(\tau)$ is also tuned among several values $[1, 0.9, 0.8, 0.7, 0.6, 0.5]n$ on the validation data. To make fair comparison, we use the same stopping criterion for all algorithms, i.e., algorithms stop if the duality gap is less than threshold $\varepsilon = 10^{-2}$ or the maximum number of iterations $T = 1000$ is reached. We repeat each experiment five times, and report the results by averaging over the 5 trials.

The left panels of Figure 1 show the classification accuracy of algorithms with noise level $q$ varied from 0 to 0.4 on the five data sets. We observe that for almost all cases, (i) the proposed algorithm is significantly more resilient than **SiPMKL** algorithm to the label noise; (ii) the worst case analysis (**StPMKL**) is better than the best case analysis (**MiPMKL**) for noisy labels, particularly when the noise level is high; and (iii) when no noise is added ($q=0$), **StPMKL** achieves similar performance, if not the same, to **SiPMKL**, while **MiPMKL** could give different results (e.g. on ionosphere, heart, sonar). The reason is that the objective in **StPMKL** reduces to the objective in **SiPMKL** when $q = 0$, however, it is not the case for **MiPMKL**. This observation is consistent with our discussion in section 3.2 above Theorem 1.

---

[2] We do not compare many other MKL algorithms because (i) some algorithms (Xu et al., 2010) optimize the same objective as in (2), (ii) some algorithms (Cortes et al., 2009; Kloft et al., 2009) focus on different regularizations (e.g. $\ell_2$ or $\ell_p$), and (iii) some algorithms (Cristianini et al., 2002; Cortes et al., 2010a) use different criteria (e.g. kernel alignment).

Table 1. Statistics of Data Sets

| Data Set | #Examples | #Features |
|---|---|---|
| ionosphere | 351 | 34 |
| heart | 270 | 13 |
| sonar | 208 | 60 |
| breast-cancer | 683 | 10 |
| australia | 690 | 14 |

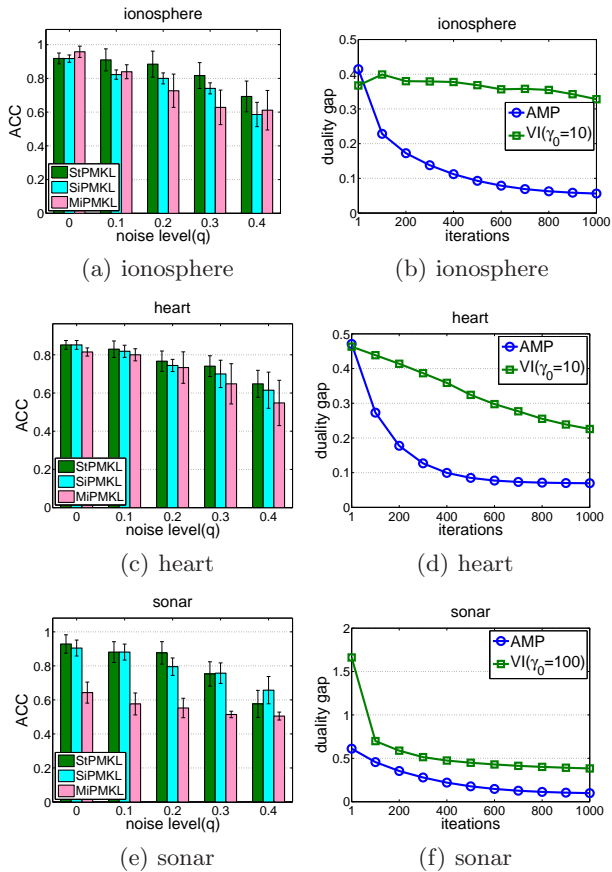

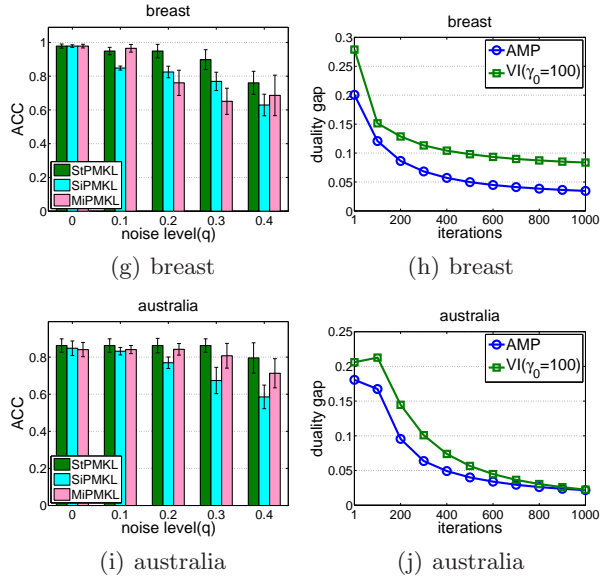

Figure 1. Comparison of accuracy (ACC) with varied noise levels (left panel) and comparison of convergence speed (right panel).

Finally, we verify the efficiency of the accelerated mirror prox method. We compare the proposed accelerated mirror prox (**AMP**) method to the variational inequality method (**VI**) (Nemirovski, 1994) (i.e. gradient descent method for convex-concave problem). For fair comparison, we fix $\lambda = 0.01$ and $\rho = 100$. The step size in variational inequality method is set to $\gamma_0/\sqrt{T}$ where we tune $\gamma_0$ in the range of $[0.01, 0.1, 1, 10, 100]$ and the best convergence with the best $\gamma_0$ is finally reported. We run both algorithms with 1000 iterations, and plot the duality gap versus the number of iterations. The results are shown in the right panels of Figure 1, which verify that the accelerated mirror prox method is significantly more efficient than the variational inequality method.

## 5. Conclusions

In this paper, we present a stochastic programming framework for multiple kernel learning from noisy labels. We formulate the problem into a convex-concave optimization problem. We also present an efficient accelerated mirror prox method for solving the related convex-convex problem. Empirical studies in a simulated environment verify the effectiveness of the proposed framework and the efficiency of the developed optimization algorithm. For future work, we plan to apply the proposed approach to real-world problems with inherent noise in labels where the noise may not be synthetically generated by independent random flipping. An open problem associated with it would be how to obtain the knowledge of the noise level.

# Acknowledgement

This work is in part supported by National Science Foundation (IIS-0643494), Office of Navy Research (ONR Award N00014-09-1-0663 and N00014-12-1-0431).